\pdfoutput=1

\documentclass[11pt]{article}

\usepackage{EMNLP2023}

\usepackage{times}
\usepackage{latexsym}
\usepackage{graphicx,tabularx}
\usepackage{amsmath,amsfonts,amssymb,amsthm,mathrsfs}
\usepackage{tikz}
\usepackage{pgfplots}
\usepackage{pgfplots}
\pgfplotsset{width=\linewidth,compat=1.9}
\usepackage{algpseudocode}
\usepackage{algorithm}
\usepackage[frozencache, cachedir=_minted-output/]{minted}

\usepackage[utf8]{inputenc}

\usepackage{microtype}

\usepackage{inconsolata}

\usepackage{multirow}

\newcommand{\MS}{\textit{{M-cont}}}
\newcommand{\MD}{{\textit{M-disc}}}
\newcommand{\Hu}{{\textit{human}}}

\newcommand{\optS}{{OPT-350m}}
\newcommand{\optL}{{OPT-1.3b}}

\newcommand{\overlap}{{activation overlap}}

\title{Unnatural language processing:\\
How do language models handle machine-generated prompts?}

\author{Corentin Kervadec \and Francesca Franzon \\
        Universitat Pompeu Fabra (UPF) / Barcelona \\
        \texttt{\{name.lastname\}@upf.edu}
        \\\And
        Marco Baroni \\
        UPF and ICREA / Barcelona\\
        \texttt{marco.baroni@upf.edu} \\}

\begin{document}
\maketitle
\begin{abstract}
Language model prompt optimization research has shown that semantically and grammatically well-formed manually crafted prompts are routinely outperformed by automatically generated token sequences with no apparent meaning or syntactic structure, including sequences of vectors from a model's embedding space. We use machine-generated prompts to probe how models respond to input that is not composed of natural language expressions. We study the behavior of models of different sizes in multiple semantic tasks in response to both continuous and discrete machine-generated prompts, and compare it to the behavior in response to human-generated natural-language prompts. Even when producing a similar output, machine-generated and human prompts trigger different response patterns through the network processing pathways, including different perplexities, different attention and output entropy distributions, and different unit activation profiles. 
We provide preliminary insight into the nature of the units activated by different prompt types, suggesting that only natural language prompts recruit a genuinely linguistic circuit.
\end{abstract}

\section{Introduction}
\label{sec:introduction}

Neural language models (LMs) are parameterized probabilistic models that can assign a probability to any sequence of language tokens. Given that they are trained on huge amounts of natural language, we expect their statistics to mimic those of the latter. In this paper, we study what happens when a LM trained on English must process ``unnatural language'', that is, sequences that are extremely unlikely in English, as they are syntactically and semantically ill-formed.

We tackle this topic through the lens of \textit{machine-generated prompts}, that is, automatically discovered input token sequences that optimize the model's performance in a target zero-shot task \cite{Shin:etal:2020,Deng:etal:2022}. It has indeed been widely observed that such prompts, while empirically effective, consist of nonsensical sequences of jumbled tokens. For example, using the popular AutoPrompt algorithm of \citet{Shin:etal:2020} and the \optL{} language model \cite{Zhang:etal:2022b}, we found that the prompt ``\textit{[X] Antarctica = sequelsStationrough [Y]}'' outperforms reasonable human-crafted prompts such as ``\textit{[X] belongs to the continent of [Y]}'' on the task of retrieving the continent a geographic body belongs to. Even more extremely, recent prompt generation methods find  sequences of embedding vectors that do not correspond to items in the model vocabulary, but still outperform both human-crafted and machine-derived discrete prompts \cite{Lester:etal:2021,liu2023pre, Zhong:etal:2021}. This state of affairs is paradoxical: why does a LM that has been trained to reproduce the statistics of natural language respond better to input sequences that are completely outside this distribution?

We present a detailed comparative study of how LMs internally process manually-crafted prompts and both discrete and continuous machine-generated prompts. While we do not solve the puzzle of why linguistically ill-formed machine-generated prompts are better than human prompts, we discover that there are fairly deep differences characterizing the various prompt types through all the processing stages of a LM, suggesting that the latter has fortuitously developed a distinct pathway to process unnatural language.

\section{Related work}
\label{sec:related}

\paragraph{Understanding prompts.} The advent of zero-shot prompting stimulated interest in the linguistic and semantic properties of prompts.\footnote{There is also related work on the effect of ablations such as word order permutations in the context of models fine-tuned for a specific task, such as natural language inference \citep[e.g.,][]{Gupta:etal:2021,Pham:etal:2021,Sinha:etal:2021,Sinha:etal:2021b}.} For example, \citet{Webson:Pavlick:2022} showed that, with minimal fine-tuning, highly semantically irrelevant prompts can be as effective as prompts with pertinent semantic content. Starting with \citet{Wallace:etal:2019} and \citet{Shin:etal:2020}, the fact that inscrutable machine-generated discrete prompts outperform natural language sequences has also attracted attention. For example, \citet{Deng:etal:2022} showed that constraining machine-generated prompts to be more ``language-like'' harms performance. \citet{Ishibashi:etal:2023} and \citet{Rakotonirina:etal:2023} studied how various ablations affect the performance of machine-generated prompts. The second study also demonstrated that it is possible to find discrete machine-generated prompts that are effective across a range of LMs. \citet{Khashabi:etal:2022} found that continuous prompts can be optimized to be near \textit{any} arbitrary text in embedding space, while being equally effective. These studies focus on properties of the prompts themselves. We complement them with an analysis of how LMs respond when exposed to these prompts. 

\paragraph{Understanding LMs}
More generally, understanding how LMs process unnatural linguistic input contributes to our understanding of their inner workings. Therefore, our study is also related to work on \textit{interpretability}~\cite{lipton2018mythos}, defined as the analysis of a trained model's decision policy. 
In particular, one can approach neural network interpretability by adopting a \textit{mechanistic} paradigm, consisting in directly studying the weights and their activation in order to reverse-engineer the neural network. Successful mechanistic insights have been obtained in computer vision \cite{voss:etal:2021, olah2020zoom}. \citet{cammarata:etal:2020} is an example of mechanistic interpretability applied to Tranformer LMs. In this context, the Transformer feed-forward layers have been shown to behave like key-value memories \cite{Geva:etal:2021}. Notably, as shown in \citet{dai:etal:2022}, these memory slots, also called \textit{knowledge neurons}, encode specific concepts acquired during pre-training. Even more interesting, manually editing these memories allows to causally control the prediction output \cite{Meng:etal:2022b}, suggesting that they play a central role in language processing \citep[see also][]{geva:etal:2022}. In the present paper, we show that unnatural language processing is achieved by recruiting different knowledge neurons than the ones used for natural language processing.

\section{Setup}
\label{sec:setup}

\subsection{Language model and tasks}

\paragraph{OPT family LMs}
We conduct our analyses on \optS{} and \optL{} \cite{Zhang:etal:2022b}, two pre-trained auto-regressive Transformer-based models trained on The Pile corpus \cite{pile}, whose pre-trained weights are publicly available from \href{https://huggingface.co/}{HuggingFace}. We choose auto-regressive models since LM development has increasingly shifted to this class, and OPT models since, in informal experiments, we found them to perform better on our tasks than comparable auto-regressive models available from HuggingFace (e.g., the GTP2 family). OPT models use a vocabulary set composed of 50,265 items.

\paragraph{Knowledge-retrieval tasks}
We base our experiment on the LAMA dataset~\cite{Petroni:etal:2019}. Initially designed to probe factual knowledge and commonsense in LM, this dataset is a collection of $\langle r, s, o\rangle$ triplets describing a relation $r$ between a subject $s$ and an object $o$, e.g.: $\langle$\textit{continent of}, Lavoisier Island, Antarctica$\rangle$. In particular, we use the TREx~\cite{ElSahar2018TRExAL} subset, whose test set contains 41 relations, each with up to 1,000 tuples. All the machine-generated prompts are trained using the data collected by \citet{Shin:etal:2020}, also containing 1,000 tuples per relation. Each relation defines a different knowledge retrieval task. We focused on these tasks because they require semantically contentful prompts (e.g., for the relation above, the prompt must carry some geographic information), as opposed to other setups where a prompt might simply have to describe the task at a meta-linguistic level (``\textit{translate the following sentence into Chinese}''; ``\textit{does paragraph X entail paragraph Y?}'', etc.).

\subsection{Prompts}
\paragraph{Terminology} We refer to different methods to derive prompts as \textit{prompt types}. We refer to the actual token sequences generated by a method for a certain task as \textit{templates}.

\paragraph{Human prompts}
Human prompts (\textit{\Hu{}}) come from an augmented version of \textsc{ParaRel}~\cite{Elazar:etal:2021}. \textsc{ParaRel} provides a set of near-paraphrase templates capturing each LAMA relation, e.g. ``\textit{[X] belongs to the continent of [Y]}''. ParaRel enlarged the initial templates provided by LAMA using paraphrases from LPAQA~\cite{jiang:etal:2020} and additional patterns mined from Wikipedia. Each prompt was then evaluated by a set of human experts. We further manually augmented the set with more paraphrases, and we cleaned the prompt set, e.g., by removing templates not adapted to auto-regressive LMs.

\paragraph{Machine-generated prompts}
We compare \Hu{} prompts with both discrete (\textit{\MD{}}) and continuous (\textit{\MS{}}) machine-generated prompts. The discrete ones are obtained using the popular Autoprompt~\cite{Shin:etal:2020} algorithm. For a given task, this algorithm generates a sequence of $N$ tokens relying on a gradient-guided search in the discrete LM's vocabulary space. We set template length to $N=5$, as it is the average human prompt length.
The continuous machine generated prompts are obtained using Optiprompt~\cite{Zhong:etal:2021}. For each task, Optiprompt generates a sequence of $N$ continuous vectors through optimization in the LM's embedding space. Similarly to Autoprompt, we set $N=5$. Machine-generated prompts are extracted using the LAMA-TREx training set (see above). 10 templates are obtained for each task by initializing training with different random seeds.

\paragraph{Template filtering}
We only use tasks for which we have, for each prompt type, at least one template reaching $>10\%$ accuracy. We end up with 5.9 \Hu, 8.3 \MD, and 9.0 \MS{} templates on average per task (across 21 tasks) for \optS{}, and 6.3 \Hu, 8.9 \MD, and 10 \MS{} templates on average per task (across 24 tasks) for \optL{}.\footnote{We attach the filtered template list as a supplementary archive.}

\subsection{Diagnostic metrics}

\paragraph{Accuracy}
We measure the effectiveness of a prompt type (\Hu{}, \MD{} or \MS{}) by computing its micro-accuracy (following \citet{Zhong:etal:2021}), defined as the proportion of cases where the prompted LM succesfully assigned maximum completion probability to the ground-truth object. We average across templates and LAMA tasks. It is worth noting that, contrary to other works, we did not perform any filtering on the LM's output.

\paragraph{Input perplexity and output entropy} We measure the average perplexity for each prompt type, defined as the exponentiated average negative log-likelihood of a \texttt{``[subject] [template]''} sequence, averaged across subjects, relations and templates. To characterize the LM probability distribution output, we also measure the average Shannon entropy of the output probability vector computed across all samples of the evaluation set.

\paragraph{Attention distribution}
We quantify how attention is distributed over input tokens following \citet{ramsauer:etal:2021}. For each attention head of each layer, we compute the average minimal number of attention values required to get a cumulative softmax probability mass of $0.90$. 
This value ranges from 0\% to 100\%. Intuitively, given a row of an attention map of a transformer layer, it corresponds to the number (in \%) of attention values you have to sum to reach 90\% of the total attention. Because attention values are normalized, if the attention is flat then the score will be 90\%. In contrast, if all the attention is focused on one token, then the score will be close to 1. This score decreases as the attention distribution becomes more peaky.

\paragraph{Knowledge neuron \overlap{}}
Motivated by \citet{Geva:etal:2021} and \citet{dai:etal:2022}, who empirically demonstrated that Transformer feed-forward (FF) layers act as key-value memories, or \textit{knowledge neurons}, we measure the \overlap{} of the intermediate FF units (corresponding to memory keys) between different prompt types. 
More formally, for a given Transformer layer, let $x\in\mathbb{R}^d$ be the token-wise hidden representation contextualised by the self-attention operation. The FF layer can be expressed as:
\begin{equation}
    \label{eq:unit}
    u = f(x \cdot K^T +b_K) 
\end{equation}
\begin{equation}
    FF(x) = u \cdot V +b_V
\end{equation}
\noindent{}where $V,K\in\mathbb{R}^{d\times d_m}$ are the FF parameters, $b_K, b_V$ their respective biases, and $f(\cdot)$ the non-linearity. $K$ can be seen as a set of $d_m$ keys, giving access to $d_m$ ``memory slots'' stored in $V$. In order to quantify which knowledge neurons are being accessed during prompt processing, we look at the units $u\in\mathbb{R}^{d_m}$ corresponding to the weights associated to each key-value pair (Eq.~\ref{eq:unit}).
A high overlap means that the prompts activate the same knowledge neurons, indicating similar processing by the LM. On the opposite, a low overlap suggests that the prompts trigger different activation pathways in the LM. The measure is described in more detail in Appendix \ref{sm:overlap}.

\paragraph{Input similarity}
For each pair of same-task templates, we measure the cosine similarity of their embedded representations. We then compute the average to get similarities at the prompt-type level.

\paragraph{Output agreement}
For each pair of same-task templates, we measure the proportion of test cases where the templates lead to the same prediction. We then compute prompt-type-level averages.

\begin{figure}[tb]
    \centering
    \includegraphics[width=\linewidth]{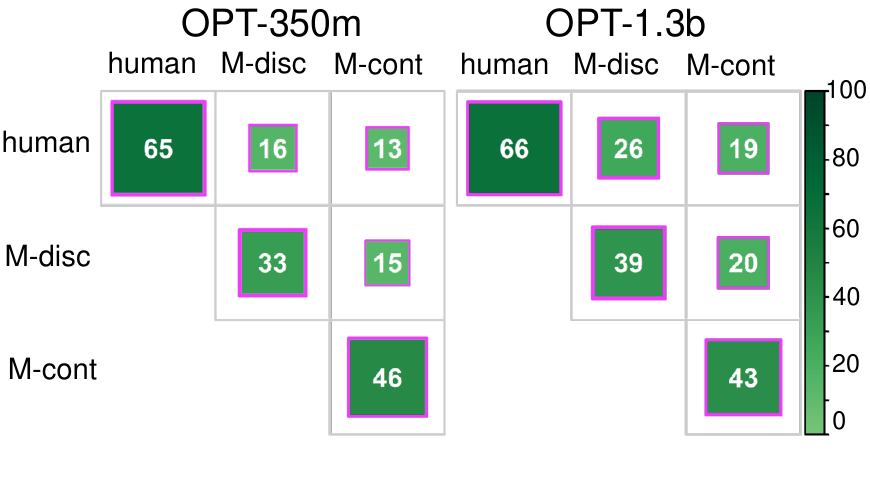}
    \caption{ Unit \overlap{} (0 to 100) between \Hu{}, \MD{} and \MS{} prompt types for \optS{} (left) and \optL{} (right). Higher values (more intense color, larger squares) represent larger overlap. Confidence intervals (CIs) are shown as square outlines: thicker lines indicate wider CIs (CIs are generally small). Within-prompt overlap is higher than betweem-types overlap, suggesting a difference in processing. }
    \label{fig:overlap}
\end{figure}

\paragraph{Uncertainty quantification}
We provide the uncertainty estimation of our measurements by computing the $95\%$ Confidence Interval (CI) of each measure. In Table~\ref{tab:ppl}, the CI associated to each metric (accuracy, perplexity, attention distribution score and output entropy) is obtained by computing the $0.025$ and $0.975$ quantiles given the list of scores obtained with each templates of a given prompt type (note that each template's score is averaged at the level of the relation). $95\%$CI in Figure~\ref{fig:overlap},\ref{fig:input-sim},\ref{fig:activation-profile}, and Table~\ref{tab:corr} are obtained using bootstrapping by randomly sampling with replacement from the list of templates (the number of resamples is found by incrementally increasing it until the uncertainty estimation converge).

\section{Processing machine-generated prompts}
\label{sec:part1}

We experimentally demonstrate that differences between human and machine-generated prompts exist at three different levels: (1) at the input level, when comparing prompt types in the embedding space, (2) at the output level, when analyzing predictions and output probabilities, and (3) at the level of intermediate activation, indicating a difference in processing at work in the LM. We conclude this quantitative analysis by showing that, although these metrics are correlated when compared within the same prompt type, the correlation is weak between prompts of different types, leading to a number of counterintuitive patterns in LM prompt processing.

\begin{table*}[tb]
\centering
\begin{tabular}{c|ccc|ccc}
\hline
\multirow{2}{*}{ } & \multicolumn{3}{c}{OPT-350m} & \multicolumn{3}{c}{OPT-1.3b}\\
& \Hu & \MD & \MS & \Hu & \MD & \MS \\
\hline

Accuracy & 29.5  & 43.4  & 54.9 & 28.8 & 46.1 & 58.0\\
\small[95\% CI] & \small[11.5, 65.0] & \small[17.0, 79.5] & \small[20.7, 86.0] & \small[10.4, 78.2] & \small[15.1, 83.4] & \small[23.8, 89.6]\\
\hline

Perplexity ($10^3$) & 0.60& 40.9 & - & 0.40 & 30.3 & - \\
\small[95\% CI] & \small[0.1, 1.9]& \small[16.0, 95.0] & & \small[0.04, 1.48] & \small[2.0, 911.3]& \\
\hline

Attention distribution (\%) & 34.4 & 30.0 & 23.2 & 30.8 & 28.7& 29.4\\
 \small[95\% CI] & \small[29.2, 39.7] & \small[27.3, 32.5] & \small[21.1, 25.5] &\small[17.0, 85.8] & \small[16.6, 84.4]& \small[14.3, 74.7]\\
\hline

Output entropy & 5.00 & 4.30& 2.10 & 4.70 & 3.90 & 2.10\\
\small[95\% CI] & \small[3.2, 6.0] & \small[1.9, 5.7]& \small[0.5, 4.3] & \small[1.7, 6.0] & \small[1.3, 6.4]& \small[0.4, 5.9]\\
 
\hline
\end{tabular}
\caption{Human and machine-generated prompts (both \MD{} and \MS) significantly differ in at least four aspects: (1) machine-generated prompts outperform the human ones in terms of accuracy; (2) they are also better calibrated on average, given their lower output entropy; while at the same time (3) machine-generated prompts are less predictable by the LMs, reaching significantly higher perplexity; (4) in machine-generated prompts, attention is concentrated on a smaller amount of tokens. For technical convenience, perplexity is not computed for \MS{}. }
\label{tab:ppl}
\end{table*}

\subsection{Human and machine-generated prompts are processed differently.}
\label{sec:different-processing}

\begin{figure}[tb]
    \centering
    \includegraphics[width=\linewidth]{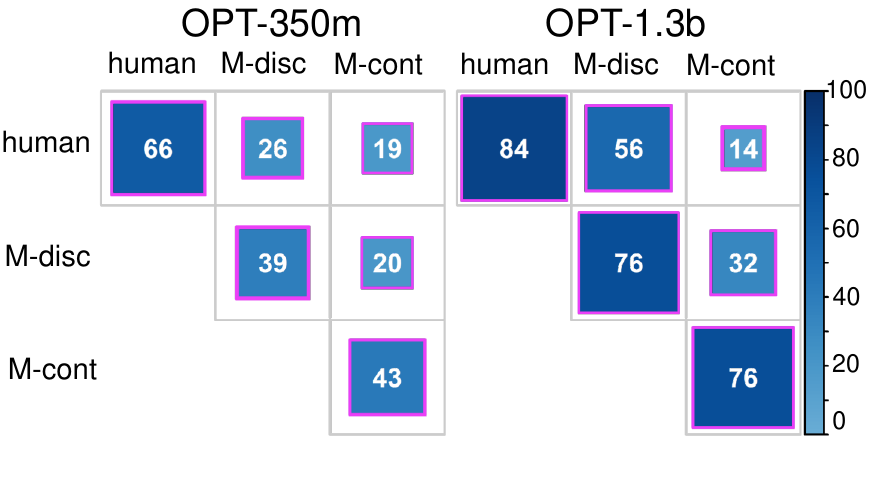}
    \caption{Percentage input similarity between \Hu{}, \MD{} and \MS{} prompt types for \optS{} (left) and \optL{} (right). Higher values (more intense color, larger squares) cue high similarity. Within-prompt-type similarities (the scores on the diagonal) are generally higher than similarity between types. Note that the absolute values of the input similarity obtained with both model sizes are not directly comparable due to a difference in the input dimension (2048 \textit{vs.} 1024).}
    \label{fig:input-sim}
\end{figure}

\begin{table*}[tb]
\centering
\begin{tabular}{r|c|c|c}
\hline
 & {Input predicts Output?} & {Input predicts Activation?} & {Activation predicts Output?}  \\
\hline
\multicolumn{4}{c}{\optS{}} \\
\hline
\Hu{} vs. \MD{}& \textbf{0.11} \small[-0.06, 0.27] & \textbf{0.21} \small[0.16, 0.26]& 0.01 \small[-0.05, 0.07] \\
\Hu{} vs. \MS{}& \textbf{0.14} \small[0.01, 0.26] & \textbf{0.06} \small[0.02, 0.09] & \textbf{-0.04} \small[-0.08, -0.00]    \\
\MD{} vs. \MS{}& \textbf{0.30} \small[0.18, 0.42] & \textbf{0.06} \small[0.03, 0.08] & \textbf{0.02} \small[-0.01, 0.05]       \\
\hline\hline
\Hu{} vs. \Hu{} & \textbf{0.53} \small[0.43, 0.62]& \textbf{0.73} \small[0.69, 0.77] & \textbf{0.66} \small[0.59, 0.72]        \\
\MD{} vs. \MD{} & \textbf{0.54} \small[0.45, 0.63]& \textbf{0.85} \small[0.83, 0.88] & \textbf{0.55} \small[0.49, 0.63]        \\
\MS{} vs. \MS{} & \textbf{0.63} \small[0.55, 0.72]& \textbf{0.74} \small[0.69, 0.78]& \textbf{0.54} \small[0.47, 0.61]         \\
\hline
\multicolumn{4}{c}{\optL{}} \\
\hline
\Hu{} vs. \MD{} & \textbf{0.18} \small[0.07, 0.28]  & \textbf{0.24} \small[0.18, 0.30]& \textbf{0.13} \small[0.05, 0.22]      \\
\Hu{} vs. \MS{} & \textbf{-0.06} \small[-0.21, 0.08]& \textbf{0.08} \small[0.03, 0.14] & \textbf{-0.03} \small[-0.08, 0.03]    \\
\MD{} vs. \MS{} & \textbf{0.12} \small[0.01, 0.22] & \textbf{0.03} \small[-0.01, 0.07] & {-0.00} \small[-0.05, 0.04]           \\
\hline\hline
\Hu{} vs. \Hu{} & \textbf{0.58} \small[0.52, 0.66]& \textbf{0.65} \small[0.61, 0.68] & \textbf{0.60} \small[0.55, 0.65]        \\
\MD{} vs. \MD{} & \textbf{0.64} \small[0.58, 0.70]& \textbf{0.89} \small[0.87, 0.90] & \textbf{0.61} \small[0.55, 0.67]        \\
\MS{} vs. \MS{} & \textbf{0.78} \small[0.73, 0.83]& \textbf{0.74} \small[0.71, 0.77]& \textbf{0.53} \small[0.49, 0.57]         \\
\hline
\multicolumn{4}{c}{\textit{Notation}: \textbf{average} \small{[95\% confidence interval]}} \\
\hline
\end{tabular}
\caption{ Pearson correlations between input similarity, output agreement and \overlap. First, we compute a single comparative statistic (input similarity,  output agreement or activation overlap) for each pair of prompts in some comparison set (e.g., \Hu{} vs.~\MD{} or \Hu{} vs.~\Hu{}); then, for each comparison set, we look at the correlation across prompt pairs between two statistics (e.g., input similarity vs.~activation overlap). Within-type correlations range from mild to high. Between-type correlations are significantly lower. These low correlations highlight counteractive aspects of LM language processing. Results in \textbf{bold} are significant (p $<0.01$). We provide the average and [95\%CI interval] correlations obtained using bootstrapped uncertainty estimation.}
\label{tab:corr}
\end{table*}

\paragraph{High accuracy and high perplexity}
As confirmed in Table~\ref{tab:ppl}, the main motivation to use machine-generated prompts is their good performance, \MS{} prompts outperforming \Hu{} ones by $+25$pts. This higher accuracy comes along with lower output entropy, suggesting better LM calibration, where a larger mass of the output probability distribution is concentrated on the correct token.\footnote{Calibration in LM analysis \citep[e.g.,][]{Liang:etal:2023} refers to the confidence that a model has in its predictions when the latter are correct. Our output entropy measure does not directly correlate confidence and accuracy. However, as machine prompts are in general more likely to trigger the correct output and, at the same time, they have lower output entropy, the global trends do suggest that they tend to produce correct answers with more confidence. We informally use the term ``calibration'' to refer to this property.} However, prompt perplexity -- quantifying the degree of predictability of a token sequence given an LM -- is two order of magnitude higher for \MD{} than for \Hu{} templates.\footnote{Due to their continuous nature, there is not trivial way to estimate perplexity for \MS{} prompts.} We discuss this further in Section \ref{sec:surprising-lm} below.

\paragraph{Low activation overlap of knowledge neurons}
Activation overlap statistics are provided in Figure~\ref{fig:overlap}. For both \optS{} and \optL{}, we observe that, while within-prompt-type overlap is mild or high, ranging from 33 to 66 (on a 0-to-100 scale), between-prompt-type overlap is always low, ranging from 13 to 26. This pattern is more pronounced when comparing \Hu{} and \MS{}. Between-prompt overlap tends to be higher with \optL{}, suggesting that larger LMs could show a convergence of human and machine-generated prompts (this remains to be further explored). 
The low-overlap result is confirmed by the diagnostic classifier analysis presented in App.~\ref{sm:diagnostic-classifiers}, that shows that a simple linear classifier can distinguish between any prompt type pair based on activation patterns on any layer of either LM.

\paragraph{Attention is focused on fewer tokens}
As transformer behaviour is a by-product of both FF and attention layers, we also look at the difference in attention distributions, shown in Table~\ref{tab:ppl}. Here again, we observe a clear distinction between human and machine-generated prompts, the latter leading to attention being focused on a smaller amount of tokens. Recall that prompt length is a hyperparameter of automated prompt induction algorithms, fixed at 5 tokens without tuning. This result might suggest that the algorithms only associated meaningful information to a subset of these tokens.

\paragraph{Machine prompts are drifting away from Human prompts in the input space.}
Figure~\ref{fig:input-sim} shows that, for both \optS{} and \optL{}, input similarity within prompt types is higher than similarity between different prompt types. In particular, the input similarity between \Hu{} and Machine prompts dramatically decreases when moving from discrete to continuous prompts.

\subsection{Surprising aspects of LM processing}
\label{sec:surprising-lm}
The significant differences that emerged between human and machine prompt processing suggest that these prompt types trigger different decision pathways. Furthermore, they provide interesting insights concerning the nature of LM processing, and, in particular, how it can occasionally be quite counter-intuitive. We explore this by considering some unexpected correlation patterns.

\paragraph{Perplexity does not predict accuracy}
\citet{Gonen:etal:2022} reported a negative correlation between perplexity and effectiveness of handcrafted prompts. However, we observe that, when using machine-generated prompt, it is possible to reach a higher prediction accuracy while having a higher perplexity. Thus, counter-intuitively, perplexity does not necessarily predict effectiveness.

\paragraph{Input similarity does not predict output agreement} The \textit{Input predicts Output?} column of Table \ref{tab:corr} measures the correlation between embedding-space similarities of same-task templates (e.g., a \Hu{} and a \MD{} template for the \textit{continent of} relation) and the rate of output agreement (defined as the portion of times different templates lead to the same prediction) for the corresponding templates. There is a significantly lower correlation when templates belonging to different prompt types are compared, especially when comparing human vs.~machine-generated templates (e.g., \Hu{} vs.~\MD{} templates), than for within-type comparisons (e.g., different \MD{} templates for the same task). When comparing different prompt types, counter-intuitively, the degree of similarity is not a good predictor of whether the templates will trigger the same output or not.

\paragraph{Activation overlap is only weakly correlated with output agreement and input similarity}
Table~\ref{tab:corr} also provides correlations between \overlap{} and input similarity (\textit{Input predicts Activation?}) or output agreement (\textit{Activation predicts Output?})~across different pairwise prompt type combinations. Within a prompt type, these correlations are mild or high, with the higher correlations pertaining to input similarity. On the contrary, \overlap{} ceases to be correlated to either input similarity or output agreement as soon as we compare different prompt types. This drop in correlation highlights the complexity of LM internal processing. Without any prior on input type, it is difficult to predict the decision pathway that will be used by the model, even in the presence of high input or output similarities.

\section{A closer look at the typical units of each prompt type}
\label{sec:part2}

\begin{figure*}[tb]
    \centering
    \includegraphics[trim=2.7cm 11cm 2.7cm 11cm, width = \linewidth]{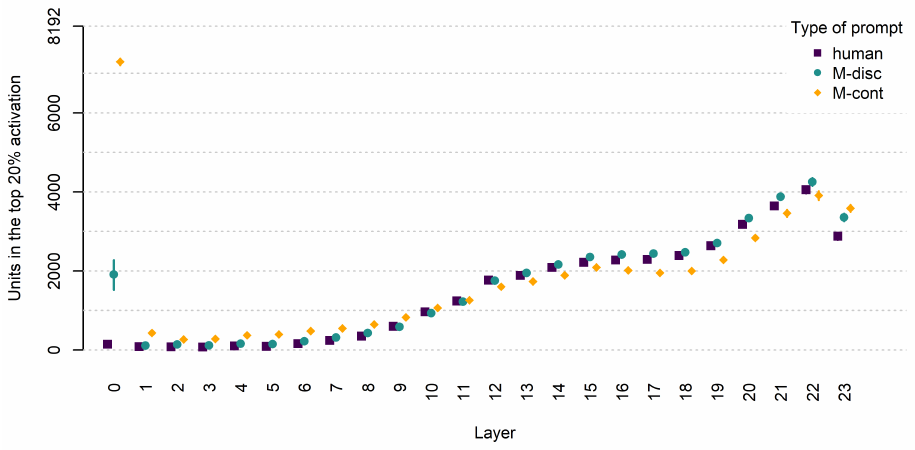}
    \caption{For each prompt type, we plot the number of units belonging to the top 20\% most activated units (overall across prompt types). \MS{} and \MD{} have significantly more highly activated units than \Hu{} on the first layer, with the effect particularly strong for \MS{}. There is also a weaker tendency for the machine prompts to activate more units on the last layer compared to the \Hu{} ones. Data from \optL{}.}
    \label{fig:activation-profile}
\end{figure*}

\subsection{Unit distribution across layers}
The low  \overlap{} between prompt types reported in Section~\ref{sec:part1} taught us that machine-generated prompts trigger units which are distinct from the ones triggered by human prompts. The units that are most often activated by the various prompt types also appear, to some degree, to be distributed differently across layers (cf.~Figure~\ref{fig:activation-profile}). In particular, machine prompts display a tendency to activate more units on the last layer and, especially, on the first one (it is worth recalling that this is the first proper Transformer layer, and not the embedding layer). The \MD{} profile lies somewhere between \Hu{} and \MS{}, confirming the trends already observed in Section~\ref{sec:part1}.

\subsection{Profiling prompt-type-typical units through associated vocabulary items}
Having shown that the three prompt types activate different pathways through the network, we seek now some insights into the nature of the units characterizing these different pathways.

\paragraph{Methodology}
We identify those units that are both \textit{typical} of a single prompt type across tasks, and significantly impacting the network output distribution, in the sense that their gradients w.r.t~to the max output probability are in the top quartile of all network units (recall that, as usual, we focus on those units we identified as knowledge neurons). We define the typical units for prompt A as those that are among the top 20\% most frequently activated by this prompt type, while at the same time being among the bottom 20\% least frequently activated by prompt types B and C.\footnote{A unit is activated when its value is greater than 0.} This filtering procedure leaves 14 \Hu{} (resp.~6), 4 \MD{} (resp.~4) and 58 \MS{} (resp.~238) units for \optS{} (resp.~\optL{}). As a sanity check, we also repeated the analysis with laxer thresholds involving more units, and the results were similar to the ones we report here.

Next, we associate each selected unit to a set of items from the LM vocabulary that strongly trigger its activation. Using the Wikipedia corpus,\footnote{Subset ``\texttt{20220301.en}'' from \href{https://huggingface.co/datasets/wikipedia}{HuggingFace}} for each item in the vocabulary we save the average unit activation in a forward pass. We sort the resulting matrix to get, for each unit, the top 500 vocabulary items leading to the strongest activation.  We  extract both input items, recording unit activation when an item is in the input sequence, and output items, recording activation when an item is predicted by the LM. We apply lower-casing and initial-space stripping on the resulting vocabulary set. More details are provided in Appendix~\ref{sm:unit-token}. This method has been chosen for its simplicity. However, it is also noisy and sensitive to rare but ``exciting'' tokens (e.g., \textit{magikarp}, see Appendix~\ref{sm:unit-token} Table~\ref{tab:unit-token}). Improving unit-item association extraction is left to future work.

Having obtained the list of vocabulary items associated to each unit, we count how many times each vocabulary item occurrs in association with any typical unit of each prompt type (as defined at the beginning of this Methodology paragraph), obtaining 3 frequency lists, one for each prompt type. We compare the relative frequencies of each vocabulary item in each list to determine which vocabulary items are most distinctively associated to (the set of typical units of) each type. In particular, using a standard method from corpus linguistics, we compute the \textit{local Mutual Information} score \citep{Evert:2005} between each vocabulary item $v$ and each prompt type $t$:

\begin{equation*}
\mathrm{LMI} = |v,t| \log \frac{P(v,t)}{P(v)P(t)}
\end{equation*}

\noindent{}where $|v,t|$ counts the occurrences of $v$ in the $t$ list, the joint probability $P(v,t)$ is estimated based on $|v,t|$; $P(v)$ is estimated using the cumulative occurrence count of $v$ in all lists, and $P(t)$ is the total number of occurrences of any item in the $t$ list.
Table \ref{tab:lmi} reports the top-30 \textit{input} vocabulary items ranked by LMI for each prompt type and both LMs.

\begin{table*}[tb]
\begin{center}
\begin{footnotesize}
\begin{tabular}{lll|lll|lll}
\hline
    \multicolumn{3}{c|}{\textit{\Hu}}&\multicolumn{3}{c|}{\textit{\MD}}&\multicolumn{3}{c}{\textit{\MS}}\\
    \hline
\multicolumn{9}{c}{\textit{\optS{}}}\\
\hline
whats  &gazed	  &ful     &handler &(\&              &361              &\^{O}\o{}$\Omega$                 &stat            &\slash\slash\slash\slash                              \\
name   &nifty	  &darn    &expr    &avascript        &ancel            &\^{O}\o{}$\Omega$\^{O}\o{}$\Omega$&page            &pwr					           \\
why    &devs	  &freaking&iterator&cpp              &yout             &\});                              &\{\textbackslash&sts					           \\
fuck   &much	  &these   &terness &addons           &risome           &());                              &0000        &]\}						   \\
noticed&like	  &have    &hillary &\textbackslash{}-&frameworks       &...........                       &+=              &table						   \\
really &daddy	  &wanna   &filename&lication         &ithub            &crossref                          &\};             &interstit.					   \\
thats  &likes	  &what    &easy    &702              &\^{O}\o{}$\Omega$&println                           &stats           &\slash**					           \\
does   &honestly  &crappy  &disabled&502              &errors           &warn                              &));             &(\{						   \\
thing  &workaround&relent  &rc	   &601               &poons            &\})                               &crip            &\slash\slash\slash\slash\slash\slash\slash\slash\slash\\
goddamn&bothers   &been    &json    &sacrific         &inline           &-----------                       &--              &debug                                                 \\
 \hline
     \multicolumn{9}{c}{\textit{\optL{}}}\\
 \hline
undermines&severe  &conducive&\^{a}\copyright&\"{a}\^{h}\textregistered{}&attle         &\^{a}ķ\`{g}                               &leilan                                          &looph      \\
curls     &dictates&frowned  &appalling      &extreme                    &cram          &\^{a}ł\`{g}\^{a}ł\`{g}\^{a}ł&\^{a}ł\^{a}ł\^{a}ł\^{a}ł&\^{}\^{}\^{}\^{}\\
makes     &tempt   &optimized&mal            &eff                        &dop           &endif                                     &everal                                          &\^{a}\textsc{k}\^{j}\\
will      &unfold  &remain   &early          &monitor                    &egregious     &canaver                                   &archdemon                                       &xff        \\
does      &bounces &reap     &complex        &schizophren                &rep           &citiz                                     &marketable                                      &..\slash   \\
manifests &persist &stroll   &hou            &rece                       &pass          &\%\%\%\%                                  &\slash\slash                                    &0000\\
prevail   &outweigh&shines   &gres           &fail                       &insanely      &\'{o}                                     &\^{a}$\hbar$¢:                                  &aeper      \\
haunt     &haun    &hangs    &crazy          &kinda                      &\~{a}\^{h}\"{}&\^{a}ķ                                    &dilig                                           &nanto      \\
meshes    &smokes  &governs  &delay          &shitty                     &prototype     &\%\%                                      &................                                &cryst      \\
wipes     &poised  &fills    &capital        &\slash\slash\slash         &devices       &\#\#\#\#\#                                &\^{a}·\^{a}·                                    &leban      \\
\end{tabular}
\end{footnotesize}
\caption{Top 30 vocabulary items associated to each prompt type ranked by LMI. Machine-generated prompts respond to less language-like items than those triggered by human prompts. Nearly all \textit{\Hu{}} items are well-formed words. Many \textit{\MD{}} items, on the other hand, are non-English diacritics, special symbols or code-related terms. \textit{\MS} items are entirely ``non-linguistic''. Some strings have been abbreviated to fit column width.}
\label{tab:lmi}
\end{center}
\end{table*}

\paragraph{Machine prompts recruit ``non-linguistic'' units}
Looking at the \optS{} results first, nearly all characteristic \textit{\Hu{}} items are well-formed words, and include a high number of forms cuing syntactic processing, such as function elements (\textit{whats, why, does}\ldots{}), inflected verbs (\textit{noticed, gazed, liked}\ldots) and modifiers (\textit{really, much, honestly}\ldots). A remarkable amount of \textit{\MD} items are coding-related terms (\textit{handler, expr, iterator}\ldots). Numbers and punctuation sequences that could be coding-related or web-page boilerplate (\textit{(\&, \textbackslash{}-}) also appear, as well as a few regular words or word fragments (\textit{Hillary, easy, sacrific}\ldots). Finally, for \textit{\MS}, the items are entirely ``non-linguistic'', being composed of sequences of non-Latin characters or punctuation marks, as well as code fragments.

Concerning \optL{}, we observe pretty much the same patterns for \textit{\Hu{}} and \textit{\MS{}}. For \textit{\MD{}}, on the other hand, together with a number of non-English diacritics and special symbols, there is a strong increase in regular words and word fragments, although the latter still clearly differ from those associated to \Hu{} prompts, in that syntax-related items, such as function words or inflected verb forms, are largely missing. 
In line with what we observed in Figure~\ref{fig:overlap}, we thus observe a cline on which, at least for \textit{\MD{}}, the difference in processing human and machine-generating prompts decreases with model size.

We tentatively conclude that machine prompts are not only triggering different activation pathways, but that the units involved in these pathways tend to respond to less language-like items than those triggered by human prompts. Note that these units are spread across the layers of the network, so that we are not only recording low-level differences in processing the input strings or vectors.\footnote{The selected typical \Hu{} units occur in layers 4th to last of \optS{} and layers 2 to 14 of \optL{} (counting from 0). The \MD{} units range from layers 3 to 10 of \optS{} and 2 to 13 of \optL{}. The \MS{} units are the only ones where, as expected given the distribution illustrated in Figure \ref{fig:activation-profile}, a significant proportion occurs on the first (0-th) layer (about one third for \optS{} and one fourth for \optL{}), but the remaining two thirds/three fourths range from layers 2 to 21 and 1 to 10, respectively.} Moreover, the results are largely mirrored by those obtained when associating units with output instead of input vocabulary items (Table \ref{tab:lmi-out} in App.~\ref{sm:vocabulary-analysis}). 

Recall that our analysis is based on units that are not only highly typical of a prompt type across relations, but also in the top gradient quartile, suggesting that they significantly contribute to the model's output distribution. It is puzzling that units that mostly respond to coding fragments or unusual characters could lead the network to produce the correct next token in the semantic tasks we are studying. We conjecture that distributed activation from such units might nudge the network towards the right output semantic fields through connectivity pathways that fortuitously arose during network training. This is an important topic for future work.

\section{Discussion}
\label{sec:discussion}

We have studied the phenomenon of linguistically and semantically opaque machine-generated prompts from the perspective of how LMs process them, compared to human-crafted ones. Our study has important \textbf{Limitations}, that are discussed in the relevant section below. However, at least for the prompt generation methods, LMs and tasks we explored, we can draw some general conclusions.

\paragraph{More than a ``happy accident''}
Our evidence suggests that the differences between human and machine-generated prompts are not just superficial, but affect all levels of network processing, and result in the activations of qualitatively different units. Some of these units are stable across semantic tasks, suggesting that they are more generally recruited to process any ``unnatural'' input. Moreover, contrary to what one could reasonably predict, there is some evidence that machine prompts are more robust than human ones, in the sense that they achieve better output calibration.

It's unlikely that the LM has been exposed to anything like \MD{} prompts during its initial training, and definitely it could not have seen out-of-vocabulary \MS{} prompts, so we can only assume that the special pathways triggered by these prompts arose through unforeseen side effects of pre-training.\footnote{We experimentally verified that model training is necessary for effective unnatural prompts to arise. We ran both Autoprompt and Optiprompt on 3 distinct random initializations of \optL{} with the same hyperparameters as in our main experiments, and found that the resulting \MD{} prompts achieved 0\% accuracy in nearly all cases, whereas \MS{} where at best able to retrieve the majority output of a task.} However, they seem to be more than just lucky connectivity accidents exploited by specific prompts to solve specific tasks, or else it would be difficult to explain the overall low entropy of machine prompt predictions and the commonalities in the units they activate. Moreover, there is evidence that \MD{} prompts can transfer across Transformer-based LMs \citep{Rakotonirina:etal:2023,Zou:etal:2023}, suggesting that unnatural language pathways might arise from the interaction of general characteristics of the Transformer architecture with Web-derived training data that are partially shared across many current pre-trained LMs. We must defer a better understanding of the nature of these unnatural pathways to future work. In particular, we plan to zoom further in into the processing of specific templates, tracking their processing throughout the network with methods such as the vocabulary-based unit analysis of Section \ref{sec:part2}.

\paragraph{On investigating unnatural language}
We believe that investigating ``unnatural language'' as we did here \citep[see also][]{Khashabi:etal:2022,Ishibashi:etal:2023,Rakotonirina:etal:2023} should be a central concern to NLP for at least three reasons. 

First, \textit{understanding} why LMs work as well as they do, and what are their failure modes, is one of the questions with the broadest scientific and societal implications we can ask today. It would however be dangerously limiting to narrow our investigation to how LMs process \textit{natural} language only, ignoring their behaviour when presented inputs outside their training distribution.

Second, unnatural language can be exploited for negative purposes, as shown by \citet{Wallace:etal:2019} and \citet{Zou:etal:2023}, who derived apparently nonsensical prompts that could steer multiple LMs' responses towards harmful behaviour, such as generating racist language.

Finally, there is recent interest in letting LMs directly communicate with each other to jointly solve tasks or to build a community \cite{Park:etal:2023,Zeng:etal:2022}. Based on our evidence, it might be pointless to insist that LM-to-LM communication takes place in natural language, given that LMs might share information more efficiently through unnatural prompts. Conversely, if being able to decode the communication flow is important (e.g., for safety reasons), care must be taken to stop LMs from drifting into unnatural language.

\paragraph{}
 For all these reasons, we hope that our preliminary contribution will encourage our community to pay more attention to the phenomenon of unnatural language processing.

\clearpage
\newpage

\section*{Limitations}
\begin{itemize}
    \item \textbf{Main limitation:} We presented an extended study of \textit{how} two pre-trained language models process human and machine-generated inputs, but we did not provide an account of \textit{why} we are observing the processing differences we are seeing. We noticed, for example, that \MS{} prompts activate units associated to punctuation marks and special characters. We do not know, however, in which way these units contribute to retrieving the correct answer in the target semantic tasks, nor how the optimization procedure chances upon them. This is our priority for future work.
    \item Our work is limited to the OPT family of models trained on the English language, to the LAMA semantic tasks and to the AutoPrompt and OptiPrompt prompt extraction methods. A straightforward direction for future work is to extend our analysis to more models (including instruction-tuned models, as instruction tuning might have a significant impact on how models respond to unnatural input), languages, data-sets and prompt extraction algorithms.
\end{itemize}

\section*{Ethics Statement}
The advent of publicly accessible LM interfaces such as ChatGPT has heated up the debate around the broader impact of LMs. While there is a variety of possible societal issues to consider \citep{Weidinger:etal:2022}, we believe that a better understanding of how LMs process information is a crucial part of bias and harm containment. If we do not understand the models, we cannot control their behaviour, and we are exposed to intentional adversarial attacks and other forms of unintentional model misuse. The very existence of completely opaque but empirically effective machine-generated prompts is proof of how counterintuitive the behaviour of LMs can be, and of how little we understand them. We thus believe that our investigations of ``unnatural language processing'' fit well into the broader program of improving our scientific understanding  of LMs, in order to make them more predictable, controllable and, ultimately, safer.

\section*{Acknowledgements}

We thank Emmanuel Chemla, Emily Cheng,  Nathana\"{e}l Carraz Rakotonirina, Xavier Suau, the members of the UPF COLT lab, the members of the Barcelona Apple Machine Learning Research group and the participants in the EviL seminar for helpful feedback and suggestions. Our work was funded by the European Research Council (ERC) under the European Union's Horizon 2020 research and innovation programme (grant agreement No.\ 101019291). This paper reflects the authors' view only, and the ERC is not responsible for any use that may be made of the information it contains.

\bibliography{marco,custom,anthology}

\begin{thebibliography}{38}
\expandafter\ifx\csname natexlab\endcsname\relax\def\natexlab#1{#1}\fi

\bibitem[{Cammarata et~al.(2020)Cammarata, Carter, Goh, Olah, Petrov, Schubert,
  Voss, Egan, and Lim}]{cammarata:etal:2020}
Nick Cammarata, Shan Carter, Gabriel Goh, Chris Olah, Michael Petrov, Ludwig
  Schubert, Chelsea Voss, Ben Egan, and Swee~Kiat Lim. 2020.
\newblock \href {https://doi.org/10.23915/distill.00024} {Thread: Circuits}.
\newblock \emph{Distill}.
\newblock Https://distill.pub/2020/circuits.

\bibitem[{Dai et~al.(2022)Dai, Dong, Hao, Sui, Chang, and Wei}]{dai:etal:2022}
Damai Dai, Li~Dong, Yaru Hao, Zhifang Sui, Baobao Chang, and Furu Wei. 2022.
\newblock Knowledge neurons in pretrained transformers.
\newblock In \emph{Proceedings of the 60th Annual Meeting of the Association
  for Computational Linguistics (Volume 1: Long Papers)}, pages 8493--8502.

\bibitem[{Deng et~al.(2022)Deng, Wang, Hsieh, Wang, Guo, Shu, Song, Xing, and
  Hu}]{Deng:etal:2022}
Mingkai Deng, Jianyu Wang, Cheng-Ping Hsieh, Yihan Wang, Han Guo, Tianmin Shu,
  Meng Song, Eric Xing, and Zhiting Hu. 2022.
\newblock {RLP}rompt: Optimizing discrete text prompts with reinforcement
  learning.
\newblock In \emph{Proceedings of EMNLP}, pages 3369--3391, Abu Dhabi, United
  Arab Emirates.

\bibitem[{Elazar et~al.(2021)Elazar, Kassner, Ravfogel, Ravichander, Hovy,
  Sch{\"u}tze, and Goldberg}]{Elazar:etal:2021}
Yanai Elazar, Nora Kassner, Shauli Ravfogel, Abhilasha Ravichander, Eduard
  Hovy, Hinrich Sch{\"u}tze, and Yoav Goldberg. 2021.
\newblock Measuring and improving consistency in pretrained language models.
\newblock \emph{Transactions of the Association for Computational Linguistics},
  9:1012--1031.

\bibitem[{ElSahar et~al.(2018)ElSahar, Vougiouklis, Remaci, Gravier, Hare,
  Laforest, and Simperl}]{ElSahar2018TRExAL}
Hady ElSahar, Pavlos Vougiouklis, Arslen Remaci, Christophe Gravier,
  Jonathon~S. Hare, Fr{\'e}d{\'e}rique Laforest, and Elena Paslaru~Bontas
  Simperl. 2018.
\newblock T-rex: A large scale alignment of natural language with knowledge
  base triples.
\newblock In \emph{International Conference on Language Resources and
  Evaluation}.

\bibitem[{Evert(2005)}]{Evert:2005}
Stephanie Evert. 2005.
\newblock \emph{The Statistics of Word Cooccurrences}.
\newblock Ph.{D} dissertation, Stuttgart University.

\bibitem[{Gao et~al.(2020)Gao, Biderman, Black, Golding, Hoppe, Foster, Phang,
  He, Thite, Nabeshima, Presser, and Leahy}]{pile}
Leo Gao, Stella Biderman, Sid Black, Laurence Golding, Travis Hoppe, Charles
  Foster, Jason Phang, Horace He, Anish Thite, Noa Nabeshima, Shawn Presser,
  and Connor Leahy. 2020.
\newblock The {P}ile: An 800gb dataset of diverse text for language modeling.
\newblock \emph{arXiv preprint arXiv:2101.00027}.

\bibitem[{Geva et~al.(2022)Geva, Caciularu, Wang, and
  Goldberg}]{geva:etal:2022}
Mor Geva, Avi Caciularu, Kevin Wang, and Yoav Goldberg. 2022.
\newblock Transformer feed-forward layers build predictions by promoting
  concepts in the vocabulary space.
\newblock In \emph{Proceedings of the 2022 Conference on Empirical Methods in
  Natural Language Processing}, pages 30--45.

\bibitem[{Geva et~al.(2021)Geva, Schuster, Berant, and Levy}]{Geva:etal:2021}
Mor Geva, Roei Schuster, Jonathan Berant, and Omer Levy. 2021.
\newblock Transformer feed-forward layers are key-value memories.
\newblock In \emph{Proceedings of EMNLP}, pages 5484--5495, Online and Punta
  Cana, Dominican Republic.

\bibitem[{Giulianelli et~al.(2018)Giulianelli, Harding, Mohnert, Hupkes, and
  Zuidema}]{Giulianelli:etal:2018}
Mario Giulianelli, Jack Harding, Florian Mohnert, Dieuwke Hupkes, and Willem
  Zuidema. 2018.
\newblock Under the hood: Using diagnostic classifiers to investigate and
  improve how language models track agreement information.
\newblock In \emph{Proceedings of the EMNLP BlackboxNLP Workshop}, pages
  240--248, Brussels, Belgium.

\bibitem[{Gonen et~al.(2022)Gonen, Iyer, Blevins, Smith, and
  Zettlemoyer}]{Gonen:etal:2022}
Hila Gonen, Srini Iyer, Terra Blevins, Noah Smith, and Luke Zettlemoyer. 2022.
\newblock Demystifying prompts in language models via perplexity estimation.
\newblock \url{https://arxiv.org/abs/2212.04037}.

\bibitem[{Gupta et~al.(2021)Gupta, Kvernadze, and Srikumar}]{Gupta:etal:2021}
Ashim Gupta, Giorgi Kvernadze, and Vivek Srikumar. 2021.
\newblock {BERT} and family eat word salad: experiments with text
  understanding.
\newblock In \emph{Proceedings of AAAI}, pages 12946--12954, Online.

\bibitem[{Ishibashi et~al.(2023)Ishibashi, Bollegala, Sudoh, and
  Nakamura}]{Ishibashi:etal:2023}
Yoichi Ishibashi, Danushka Bollegala, Katsuhito Sudoh, and Satoshi Nakamura.
  2023.
\newblock Evaluating the robustness of discrete prompts.
\newblock In \emph{Proceedings of EACL}, pages 2373--2384, Dubrovnik, Croatia.

\bibitem[{Jiang et~al.(2020)Jiang, Xu, Araki, and Neubig}]{jiang:etal:2020}
Zhengbao Jiang, Frank~F Xu, Jun Araki, and Graham Neubig. 2020.
\newblock How can we know what language models know?
\newblock \emph{Transactions of the Association for Computational Linguistics},
  8:423--438.

\bibitem[{Khashabi et~al.(2022)Khashabi, Lyu, Min, Qin, Richardson, Welleck,
  Hajishirzi, Khot, Sabharwal, Singh, and Choi}]{Khashabi:etal:2022}
Daniel Khashabi, Xinxi Lyu, Sewon Min, Lianhui Qin, Kyle Richardson, Sean
  Welleck, Hannaneh Hajishirzi, Tushar Khot, Ashish Sabharwal, Sameer Singh,
  and Yejin Choi. 2022.
\newblock Prompt waywardness: The curious case of discretized interpretation of
  continuous prompts.
\newblock In \emph{Proceedings of NAACL}, pages 3631--3643, Seattle, WA.

\bibitem[{Lester et~al.(2021)Lester, Al-Rfou, and Constant}]{Lester:etal:2021}
Brian Lester, Rami Al-Rfou, and Noah Constant. 2021.
\newblock The power of scale for parameter-efficient prompt tuning.
\newblock In \emph{Proceedings of EMNLP}, pages 3045--3059, Punta Cana,
  Dominican Republic.

\bibitem[{Liang et~al.(2023)Liang, Bommasani, Lee, Tsipras, Soylu, Yasunaga,
  Zhang, Narayanan, Wu, Kumar, Newman, Yuan, Yan, Zhang, Cosgrove, Manning,
  R\'{e}, {Acosta-Navas}, Hudson, Zelikman, Durmus, Ladhak, Rong, Ren, Yao,
  Wang, Santhanam, Orr, Zheng, Yuksekgonul, Suzgun, Kim, Guha, Chatterji,
  Khattab, Henderson, Huang, Chi, Xie, Santurkar, Ganguli, Hashimoto, Icard,
  Zhang, Chaudhary, Wang, Li, Mai, Zhang, and Koreeda}]{Liang:etal:2023}
Percy Liang, Rishi Bommasani, Tony Lee, Dimitris Tsipras, Dilara Soylu,
  Michihiro Yasunaga, Yian Zhang, Deepak Narayanan, Yuhuai Wu, Ananya Kumar,
  Benjamin Newman, Binhang Yuan, Bobby Yan, Ce~Zhang, Christian Cosgrove,
  Christopher Manning, Christopher R\'{e}, Diana {Acosta-Navas}, Drew Hudson,
  Eric Zelikman, Esin Durmus, Faisal Ladhak, Frieda Rong, Hongyu Ren, Huaxiu
  Yao, Jue Wang, Keshav Santhanam, Laurel Orr, Lucia Zheng, Mert Yuksekgonul,
  Mirac Suzgun, Nathan Kim, Neel Guha, Niladri Chatterji, Omar Khattab, Peter
  Henderson, Qian Huang, Ryan Chi, Sang Xie, Shibani Santurkar, Surya Ganguli,
  Tatsunori Hashimoto, Thomas Icard, Tianyi Zhang, Vishrav Chaudhary, William
  Wang, Xuechen Li, Yifan Mai, Yuhui Zhang, and Yuta Koreeda. 2023.
\newblock Holistic evaluation of language models.
\newblock \url{https://arxiv.org/abs/2211.09110}.

\bibitem[{Lipton(2018)}]{lipton2018mythos}
Zachary~C Lipton. 2018.
\newblock The mythos of model interpretability.
\newblock \emph{Communications of the ACM}, 61(10):36--43.

\bibitem[{Liu et~al.(2023)Liu, Yuan, Fu, Jiang, Hayashi, and
  Neubig}]{liu2023pre}
Pengfei Liu, Weizhe Yuan, Jinlan Fu, Zhengbao Jiang, Hiroaki Hayashi, and
  Graham Neubig. 2023.
\newblock Pre-train, prompt, and predict: A systematic survey of prompting
  methods in natural language processing.
\newblock \emph{ACM Computing Surveys}, 55(9):1--35.

\bibitem[{Meng et~al.(2022)Meng, Bau, Andonian, and Belinkov}]{Meng:etal:2022b}
Kevin Meng, David Bau, Alex Andonian, and Yonatan Belinkov. 2022.
\newblock Locating and editing factual associations in {GPT}.
\newblock \emph{Advances in Neural Information Processing Systems},
  35:17359--17372.

\bibitem[{Olah et~al.(2020)Olah, Cammarata, Schubert, Goh, Petrov, and
  Carter}]{olah2020zoom}
Chris Olah, Nick Cammarata, Ludwig Schubert, Gabriel Goh, Michael Petrov, and
  Shan Carter. 2020.
\newblock \href {https://doi.org/10.23915/distill.00024.001} {Zoom in: An
  introduction to circuits}.
\newblock \emph{Distill}.
\newblock Https://distill.pub/2020/circuits/zoom-in.

\bibitem[{Park et~al.(2023)Park, {O'Brien}, Cai, {Ringel Morris}, Liang, and
  Bernstein}]{Park:etal:2023}
Joon-Sung Park, Joseph {O'Brien}, Carrie Cai, Meredith {Ringel Morris}, Percy
  Liang, and Michael Bernstein. 2023.
\newblock Generative agents: Interactive simulacra of human behavior.
\newblock \url{https://arxiv.org/abs/2304.03442}.

\bibitem[{Pedregosa et~al.(2011)Pedregosa, Varoquaux, , Gramfort, Michel,
  Thirion, Grisel, Blondel, Prettenhofer, Weiss, Dubourg, Vanderplas, Passos,
  Cournapeau, Brucher, Perrot, and Duchesnay}]{Pedregosa:etal:2011}
Fabian Pedregosa, Ga{\"{e}}l Varoquaux, , Alexandre Gramfort, Vincent Michel,
  Bertrand Thirion, Olivier Grisel, Mathieu Blondel, Peter Prettenhofer, Ron
  Weiss, Vincent Dubourg, Jake Vanderplas, Alexandre Passos, David Cournapeau,
  Matthieu Brucher, Matthieu Perrot, and {\'{E}}douard Duchesnay. 2011.
\newblock {Scikit-learn}: Machine learning in {P}ython.
\newblock \emph{Journal of Machine Learning Research}, 12:2825--2830.

\bibitem[{Petroni et~al.(2019)Petroni, Rockt{\"a}schel, Riedel, Lewis, Bakhtin,
  Wu, and Miller}]{Petroni:etal:2019}
Fabio Petroni, Tim Rockt{\"a}schel, Sebastian Riedel, Patrick Lewis, Anton
  Bakhtin, Yuxiang Wu, and Alexander Miller. 2019.
\newblock Language models as knowledge bases?
\newblock In \emph{Proceedings EMNLP}, pages 2463--2473, Hong Kong, China.

\bibitem[{Pham et~al.(2021)Pham, Bui, Mai, and Nguyen}]{Pham:etal:2021}
Thang Pham, Trung Bui, Long Mai, and Anh Nguyen. 2021.
\newblock {Out of Order}: How important is the sequential order of words in a
  sentence in natural language understanding tasks?
\newblock In \emph{Findings of ACL}, pages 1145--1160, Online.

\bibitem[{Rakotonirina et~al.(2023)Rakotonirina, Dess\`{i}, Petroni, Riedel,
  and Baroni}]{Rakotonirina:etal:2023}
Nathana\"{e}l Rakotonirina, Roberto Dess\`{i}, Fabio Petroni, Sebastian Riedel,
  and Marco Baroni. 2023.
\newblock Can discrete information extraction prompts generalize across
  language models?
\newblock In \emph{Proceedings of ICLR}, Kigali, Rwanda.
\newblock Published online:
  \url{https://openreview.net/group?id=ICLR.cc/2023/Conference}.

\bibitem[{Ramsauer et~al.(2021)Ramsauer, Sch{\"a}fl, Lehner, Seidl, Widrich,
  Gruber, Holzleitner, Adler, Kreil, Kopp et~al.}]{ramsauer:etal:2021}
Hubert Ramsauer, Bernhard Sch{\"a}fl, Johannes Lehner, Philipp Seidl, Michael
  Widrich, Lukas Gruber, Markus Holzleitner, Thomas Adler, David Kreil,
  Michael~K Kopp, et~al. 2021.
\newblock Hopfield networks is all you need.
\newblock In \emph{International Conference on Learning Representations 2021}.

\bibitem[{Shin et~al.(2020)Shin, Razeghi, Logan~IV, Wallace, and
  Singh}]{Shin:etal:2020}
Taylor Shin, Yasaman Razeghi, Robert Logan~IV, Eric Wallace, and Sameer Singh.
  2020.
\newblock {A}uto{P}rompt: {E}liciting knowledge from language models with
  automatically generated prompts.
\newblock In \emph{Proceedings of EMNLP}, pages 4222--4235, Online.

\bibitem[{Sinha et~al.(2021{\natexlab{a}})Sinha, Jia, Hupkes, Pineau, Williams,
  and Kiela}]{Sinha:etal:2021}
Koustuv Sinha, Robin Jia, Dieuwke Hupkes, Joelle Pineau, Adina Williams, and
  Douwe Kiela. 2021{\natexlab{a}}.
\newblock Masked language modeling and the distributional hypothesis: Order
  word matters pre-training for little.
\newblock In \emph{Proceedings of EMNLP}, pages 2888--2913, Punta Cana,
  Dominican Republic.

\bibitem[{Sinha et~al.(2021{\natexlab{b}})Sinha, Parthasarathi, Pineau, and
  Williams}]{Sinha:etal:2021b}
Koustuv Sinha, Prasanna Parthasarathi, Joelle Pineau, and Adina Williams.
  2021{\natexlab{b}}.
\newblock {UnNatural} {L}anguage {I}nference.
\newblock In \emph{Proceedings of ACL}, pages 7329--7346, Online.

\bibitem[{Voss et~al.(2021)Voss, Cammarata, Goh, Petrov, Schubert, Egan, Lim,
  and Olah}]{voss:etal:2021}
Chelsea Voss, Nick Cammarata, Gabriel Goh, Michael Petrov, Ludwig Schubert, Ben
  Egan, Swee~Kiat Lim, and Chris Olah. 2021.
\newblock \href {https://doi.org/10.23915/distill.00024.007} {Visualizing
  weights}.
\newblock \emph{Distill}.
\newblock Https://distill.pub/2020/circuits/visualizing-weights.

\bibitem[{Wallace et~al.(2019)Wallace, Feng, Kandpal, Gardner, and
  Singh}]{Wallace:etal:2019}
Eric Wallace, Shi Feng, Nikhil Kandpal, Matt Gardner, and Sameer Singh. 2019.
\newblock Universal adversarial triggers for attacking and analyzing {NLP}.
\newblock In \emph{Proceedings of EMNLP}, pages 2153--2162, Hong Kong, China.

\bibitem[{Webson and Pavlick(2022)}]{Webson:Pavlick:2022}
Albert Webson and Ellie Pavlick. 2022.
\newblock Do prompt-based models really understand the meaning of their
  prompts?
\newblock In \emph{Proceedings of NAACL}, pages 2300--2344, Seattle, WA.

\bibitem[{Weidinger et~al.(2022)Weidinger, Uesato, Rauh, Griffin, Huang,
  Mellor, Glaese, Cheng, Balle, Kasirzadeh, Biles, Brown, Kenton, Hawkins,
  Stepleton, Birhane, Hendricks, Rimell, Isaac, Haas, Legassick, Irving, and
  Gabriel}]{Weidinger:etal:2022}
Laura Weidinger, Jonathan Uesato, Maribeth Rauh, Conor Griffin, Po-Sen Huang,
  John Mellor, Amelia Glaese, Myra Cheng, Borja Balle, Atoosa Kasirzadeh,
  Courtney Biles, Sasha Brown, Zac Kenton, Will Hawkins, Tom Stepleton, Abeba
  Birhane, Lisa~Anne Hendricks, Laura Rimell, William Isaac, Julia Haas, Sean
  Legassick, Geoffrey Irving, and Iason Gabriel. 2022.
\newblock Taxonomy of risks posed by language models.
\newblock In \emph{Proceedings of FAccT}, pages 214--229, Seoul, Korea.

\bibitem[{Zeng et~al.(2023)Zeng, Attarian, Ichter, Choromanski, Wong, Welker,
  Tombari, Purohit, Ryoo, Sindhwani, Lee, Vanhoucke, and
  Florence}]{Zeng:etal:2022}
Andy Zeng, Maria Attarian, Brian Ichter, Krzysztof Choromanski, Adrian Wong,
  Stefan Welker, Federico Tombari, Aveek Purohit, Michael Ryoo, Vikas
  Sindhwani, Johnny Lee, Vincent Vanhoucke, and Pete Florence. 2023.
\newblock Socratic models: {Composing} zero-shot multimodal reasoning with
  language.
\newblock In \emph{Proceedings of ICLR}, Kigali, Rwanda.
\newblock Published online:
  \url{https://openreview.net/group?id=ICLR.cc/2023/Conference}.

\bibitem[{Zhang et~al.(2022)Zhang, Roller, Goyal, Artetxe, Chen, Chen, Dewan,
  Diab, Li, Lin, Mihaylov, Ott, Shleifer, Shuster, Simig, Koura, Sridhar, Wang,
  and Zettlemoyer}]{Zhang:etal:2022b}
Susan Zhang, Stephen Roller, Naman Goyal, Mikel Artetxe, Moya Chen, Shuohui
  Chen, Christopher Dewan, Mona Diab, Xian Li, Xi~Lin, Todor Mihaylov, Myle
  Ott, Sam Shleifer, Kurt Shuster, Daniel Simig, Punit Koura, Anjali Sridhar,
  Tianlu Wang, and Luke Zettlemoyer. 2022.
\newblock {OPT}: Open pre-trained transformer language models.
\newblock \url{https://arxiv.org/abs/2205.01068}.

\bibitem[{Zhong et~al.(2021)Zhong, Friedman, and Chen}]{Zhong:etal:2021}
Zexuan Zhong, Dan Friedman, and Danqi Chen. 2021.
\newblock Factual probing is [{MASK}]: Learning vs.~learning to recall.
\newblock In \emph{Proceedings of NAACL}, pages 5017--5033, Online.

\bibitem[{Zou et~al.(2023)Zou, Wang, Kolter, and Fredrikson}]{Zou:etal:2023}
Andy Zou, Zifan Wang, Zico Kolter, and Matt Fredrikson. 2023.
\newblock Universal and transferable adversarial attacks on aligned language
  models.
\newblock \url{http://arxiv.org/abs/2307.15043}.

\end{thebibliography}
\bibliographystyle{acl_natbib}

\clearpage
\newpage
\appendix

\section{Measuring the overlap of activated knowledge neurons}
\label{sm:overlap}

In Section~\ref{sec:part1}, we measure and compare which knowledge neurons,
 that is, units found in the intermediate layers of the feed-forward Transformer blocks \cite{dai:etal:2022, geva:etal:2022}, are activated by different prompt types. In particular, we measure the \textit{knowledge neuron activation overlap} (abbreviated as \overlap). The following algorithms in pseudo-code detail how this measure is obtained. 

\paragraph{}
First, we construct a Boolean matrix recording which units are activated by a template. As shown in the pseudo-code below, a unit is said to be activated if its value is greater than 0 on more than $k=20\%$ of cases when instantiated with each of the subjects associated to the template relation. In the pseudo-code, $\texttt{Relations}$ is the set of relations available in our task set; given a relation, $\texttt{Subjects}$ provides the list of relevant subjects and $\texttt{template(s)}$ instantiates the template with subject $\texttt{s}$. $\texttt{Model}$ is the LM being used.
\begin{minted}{python}
def get_act(template, relation):
    cpt = 0
    k = 0.2
    # iterate across subject
    for s in Subjects(relation):
        inpt = template(s)
        for u in Model(inpt).units:
            cpt[u] += (u>0)
    # u is activated if >0 for
    # more than k% of inputs
    act = cpt > k*len(Subjects(relation))
    return act
\end{minted}

Then, for each pair of templates (in \texttt{Templates}), we compute the intersection over union of their respective activation matrices:
\begin{minted}{python}
overlap = {}
for r in Relations:
    for t_A in Templates(r):
        for t_B in Templates(r):
            act_A = get_act(t_A,r)
            act_B = get_act(t_B,r)
            i = act_A & act_B
            u = act_A | act_B
            overlap[(t_A, t_B)] = i/u
\end{minted}

Finally, we average these pairwise overlaps while filtering by prompt type (e.g., only averaging the \overlap{} for pairs containing one \Hu{} and one \MS{}).

\begin{table}[tb]
    \centering
    \begin{tabular}{ccc}
        \hline
        \multicolumn{3}{c}{Layer 0, unit 248} \\
        \hline
        ``( & >>>>>>>> & ..............\\
        {[}\_ & .......... & ="/ \\
        :::::::: & \{: & ,,,, \\
        \hline
        \multicolumn{3}{c}{Layer 10, unit 674} \\
        \hline
        antidepress & debian & frieza \\
        magikarp & minecraft & xperia \\
        oneplus & awakens & bitcoin \\
        \hline
        \multicolumn{3}{c}{Layer 16, unit 2126} \\
        \hline
        filename & windows & drm \\
        sshd & misunder & rm \\
        folder & pkg & vm `\\
        \hline
        \multicolumn{3}{c}{Layer 22, unit 3617} \\
        \hline
        everal & huge & every \\ 
        risome & some & any \\ 
        these  & crappy & nifty \\
    \end{tabular}
    \caption{For each intermediate unit in the OPT's feed-forward layers we extract the set of items leading to the strongest activation on the Wikipedia corpus. As an illustration, we selected four different units with distinct profiles and displayed them in this table, along with their top 9 most associated input items, for \optS{}. We observe varying degrees of consistency and naturalness across units.}
    \label{tab:unit-token}
\end{table}

\section{Diagnostic classifiers}
\label{sm:diagnostic-classifiers}

To complement the activation-overlap-based analysis presented in Section \ref{sec:different-processing} of the main paper, we run a set of shallow linear ``diagnostic'' classifiers \citep{Giulianelli:etal:2018} of the activations generated by the models on each layer in response to inputs from each prompt type. As usual, we focus on the activation of knowledge neurons.

\begin{figure*}[tb]
    \centering
    \includegraphics[width = \hsize]{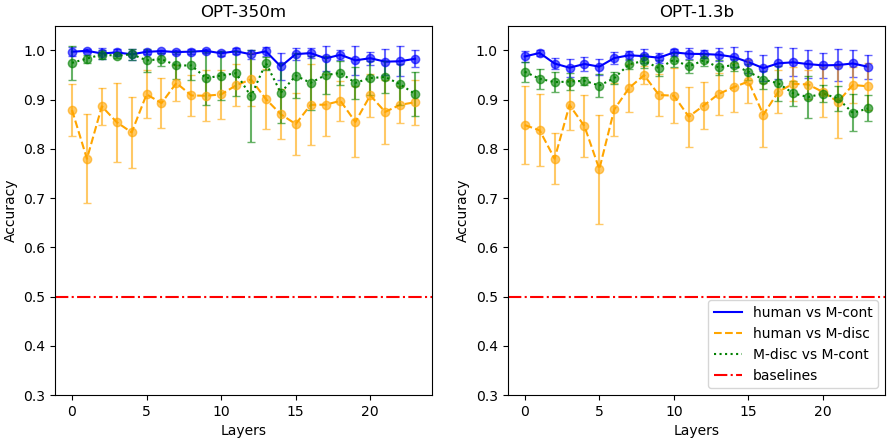}
    \caption{Average accuracies and standard deviations for 20 runs of the pairwise prompt-type classification experiments across the 24 layers of the networks.}
    \label{fig:classification_performance}
\end{figure*}

\paragraph{Data} As in the main paper, we use all templates with LAMA accuracy $>=10\%$, filtering out a random subset of the LAMA P176 relation \Hu{} templates, as this relation is greatly over-represented. We are left with 21 and 24 tasks for \optS{} and \optL{}, respectively. As we are interested in units inherently distinguishing prompt types independently of lexico-semantic aspects associated to specific templates or tasks, we partition the data so that the training and test data contain disjoint tasks (and, \textit{a fortiori}, disjoint templates). We consider 4 such partitions, each time using data from 16 (\optS)/18 (\optL) tasks for training and 5 (\optS) /6 (\optL) for testing, such that there is no test task overlap across the partitions.\footnote{As we have 21 total tasks meeting our conditions for \optS{}, one of the 4 partitions used for this model consists of 15 training and 6 test tasks} We instantiate each template with 10 randomly selected subjects from the corresponding LAMA lists. For each pairwise classification task, we balance the test instances by downsampling the larger class, so that chance/majority/minority accuracies (``baselines'' in Figure \ref{fig:classification_performance}) are at 50\%.

\paragraph{Classifier} We use a logistic regression classifier with L1 regularization (to encourage sparseness), with the L1 term coefficient fixed at  $\alpha=0.01$. We fit the classifier with stochastic gradient descent, using the Scikit-learn toolkit \citep{Pedregosa:etal:2011}. For each of the 4 training partitions, we repeat the experiment with 5 different seeds, resulting in a total of 20 runs for each layer.

\textbf{Results.} Figure \ref{fig:classification_performance} reports average per-layer classification accuracies for each pairwise prompt type comparison, with standard deviations across the 20 runs. In all cases, accuracy values are well above baseline level, and typically very high. We conclude that each single layer contains units whose activation is sufficiently discriminative for each prompt type to successfully train the classifiers, despite the challenging setup in which training and test tasks (and consequently templates) are completely disjoint. Interestingly, few units on each layer suffice to discriminate prompt types (the average  classifier weight sparsity across all experiments is at 99.5\% with 0.2\% standard deviation for \optS{} and 99.6\% with 0.4\% s.d.~for \optL). The easiest distinctions involve \MS{} as one of the classes, confirming that out-of-vocabulary embeddings make continuous prompts particularly different from natural language. Indeed, it's remarkable that distinguishing \MD{} from \MS{} is generally easier than distinguishing between \MD{} and \Hu{} prompts.

To conclude, the classification experiments bring strong convergent evidence that genuinely different pathways characterize different prompt types across all layers of the network.

\section{Unit/vocabulary item association}
\label{sm:unit-token}

\paragraph{Implementation details}
When extracting unit/vocabulary-item association, we empirically set the window size to $15$. This value is reasonable close to prompt size ($5$ in average), while containing a sufficient amount of tokens to get a meaningful context. In addition, we set the window stride to $15$ to save computation time. Furthermore, due to our limited computation resources, and given the size of the Wikipedia corpus (6B), we only used 66\% of the data for \optL{}.

\paragraph{Samples}
As illustrated in Table~\ref{tab:unit-token}, we came up with a large diversity of unit profiles, some of them being associated to more or less linguistically valid items, and with  varying degree of semantic consistency.

\section{Profiling typical units by output vocabulary analysis}
\label{sm:vocabulary-analysis}

Table \ref{tab:lmi} in the main paper reports the 30 \textit{input} vocabulary items with the largest LMI with respect to each prompt type. Table \ref{tab:lmi-out} here reports the top-30 \textit{output} items. We largely confirm the same trends, although we do notice an overall tendency for the units triggered by machine prompts to be associated to more ``language-like'' output material, which makes sense as ultimately these prompts do produce well-formed task-relevant outputs.

\begin{table*}[tb]
\begin{footnotesize}
\hskip-1.0cm\begin{tabular}{lll|lll|lll}
\hline
    \multicolumn{3}{c|}{\textit{\Hu}}&\multicolumn{3}{c|}{\textit{\MD}}&\multicolumn{3}{c}{\textit{\MS}}\\
    \hline
\multicolumn{9}{c}{\textit{\optS{}}}\\
\hline
nobody   &undone         &tonight      &incompet                      &fanc  &\{\              &\^{O}\o{}$\Omega$                 &,\~{n}\`{a}&compare            \\
yesterday&before         &scrambled    &alluded                       &ridic &\textbackslash{}'&\^{O}\o{}$\Omega$\^{O}\o{}$\Omega$&",         &pref               \\
okay     &anybody        &reasonably   &\textbackslash{}\textbackslash&782   &\"{}\%           &tracker                           &\});       &also               \\
really   &plugged        &messed       &):                            &xcom  &blat             &;;                                &moreover   &comple             \\
awoken   &captcha        &pinpoint     &juxtap                        &":["  &forge            &checking                          &['         &attempt            \\
bother   &interchangeable&right        &"\textbackslash{}             &-|    &physic           &supported                         &therefore  &'\textbackslash{});\\
corrobor &glanced        &authenticated&revert                        &698   &tyr              &meanwhile                         &text       &0000          \\
parsed   &glean          &snowball     &insin                         &(\{   &dst              &avg                               &|--        &((                 \\
bothering&tweaked        &earlier      &faintly                       &invoke&772              &insert                            &>>>        &currently          \\
nailed   &figured        &tasted       &irresist                      &mysql &ceremon          &header                            &\},"       &prev               \\
 \hline
     \multicolumn{9}{c}{\textit{\optL{}}}\\
 \hline
accumulating &snug        &peeled    &reconc       &oddly       &localhost &\^{a}k\'{g}                                                 &\^{a}$\bar{}$                 &\^{a}ł$\mathrm{\dot{g}}$\^{a}ł$\mathrm{\dot{g}}$\^{a}ł\\
elic         &attributable&distinctly&monstrous    &vaugh       &filib     &\^{a}ł\^{a}ł                                                &\^{a}·\^{a}                   &\u{o}ł                                                                        \\
overpower    &solely      &ooz       &trembling    &prett       &though    &\^{a}\'{l}ı                                                 &\~{a}\'{h} \u{\i}             &inst                                                                          \\
substantially&unmatched   &waging    &check        &outlandish  &recip     &\`{i}\P                                                     &\^{a}ł$\mathrm{\dot{g}}$\^{a}ł&\}\{                                                                          \\
fundamentally&overloaded  &incred    &adolesc      &mpg         &disag     &kinnikuman                                                  &say                           &\^{a}\~{\i}ĳ                                                                  \\
reinvest     &obligatory  &impecc    &\^{a}ī¼      &pause       &acquies   &\%\%\%\%                                                    &+=                            &\^{a}\'{l}\~{\i}                                                              \\
unparalleled &infused     &deprecated&understanding&independ    &murky     &===                                                         &any                           &*****                                                                         \\
anonym       &constrained &inherently&collaps      &bicy        &budgetary &services                                                    &sample                        &\"{\i}\c{ }\i                                                                 \\
overseen     &sandwic     &surg      &disbel       &scram       &game      &provider                                                    &\^{a}\P                       &\~{a}\^{h}\~{i}                                                               \\
ideally      &achievable  &tempted   &unpop        &foundational&billboards&\^{a}ł\^{a}ł\^{a}ł\^{a}ł\^{a}ł\^{a}ł\^{a}ł&url                           &\^{}\^{}\^{}\^{}                                                                      \\
\end{tabular}
\end{footnotesize}
\caption{Top 30 output items associated to each prompt type ranked by LMI.  Some strings have been abbreviated to fit column width.}
\label{tab:lmi-out}
\end{table*}

\end{document}